%% file: ms.tex
\documentclass[runningheads]{llncs}
\usepackage{graphicx}
\usepackage{amsmath,amssymb} 

\usepackage{color}

\usepackage{times}
\usepackage{epsfig}
\usepackage{dcolumn}
\usepackage{amsmath}
\usepackage{amssymb}

\usepackage{subfiles}
\usepackage{hyperref}

\usepackage{algorithm}
\usepackage{algorithmic}

\newcommand{\R}{\mathbb{R}}

\renewcommand{\vec}[1]{\mathbf{#1}}

\DeclareMathOperator{\vectorize}{vec}
\DeclareMathOperator*{\argmin}{arg\,min}

\graphicspath{{Figures/}}

\newcommand\redtext{\color{black}}

\begin{document}
\pagestyle{headings}
\mainmatter

\def\ACCVSubNumber{798}  

\title{Embedded polarizing filters to separate diffuse and specular reflection} 
\titlerunning{Embedded polarizing filters to separate diffuse and specular reflection}
\authorrunning{Jospin et al.}

\author{Laurent Valentin Jospin\and
Gilles Baechler\and
Adam Scholefield}
\institute{Paper ID \ACCV18SubNumber}

\institute{Ecole Polytechnique F\'{e}d\'{e}rale de Lausanne, Switzerland \\
School of Computer and Communication Sciences\\
Audiovisual Communications Laboratory\\
\email{laurent.jospin@alumni.epfl.ch}}

\maketitle

\begin{abstract}
Polarizing filters provide a powerful way to separate diffuse and specular reflection; however, traditional methods rely on several captures and require proper alignment of the filters.
Recently, camera manufacturers have proposed to embed polarizing micro-filters in front of the sensor, creating a mosaic of pixels with different polarizations. 
In this paper, we investigate the advantages of such camera designs. In particular, we consider different design patterns for the filter arrays and propose an algorithm to demosaic an image generated by such cameras. 
This essentially allows us to separate the diffuse and specular components using a single image.
The performance of our algorithm is compared with a color-based method using synthetic and real data.
Finally, we demonstrate how we can recover the normals of a scene using the diffuse images estimated by our method.

\redtext{}
\end{abstract}

\input{./sections/content}

\input{./sections/conclusion}

\pagebreak

\bibliographystyle{splncs}
\bibliography{egbib}

\end{document}

%% file: sections/content.tex

\section{Introduction}

The light reflected from a scene can be broadly classified into \emph{diffuse} and \emph{specular} terms. While the diffuse term is slowly varying for different pairs of viewing and lighting angles, specularities take the form of strong highlights that can vary quickly as either angle changes. 
Separating these two components is useful in many  different imaging applications. For example, photometric stereo~\cite{Woodham1980} estimates surface normals from images taken under different illuminations; however, like many algorithms, the process assumes that the scene is Lambertian, i.e the reflection is purely diffuse.

As a second example, consider the problem of estimating spatially-varying bidirectional reflectance distribution functions (SVBRDFs). 
The knowledge of the complete SVBRDF enables the reproduction of objects from any viewing direction and under variable lighting conditions.
Typically, one fits a parametric model to samples obtained by capturing images with different viewing and lighting directions. Although for very simple materials the specular and diffuse components of the model can be fit from only the shape of the SVBRDF, often this process can be done much more accurately if the two components are separated prior to model fitting.
Other applications that benefit from this separation include tracking, object recognition and image segmentation.

\subsection{Related work}

There are a number of approaches to perform this separation, using either a single or multiple images~\cite{Artusi2011}. Most single-image approaches are color-based: this process was pioneered by Shafer~\cite{Shafer1985}, with the so-called \emph{dichromatic reflection model}.
A key observation is that, for dielectrics, the spectrum of the specularity is similar to the spectrum of the light source, whereas the diffuse spectrum also encompasses information about the color of the surface: this results in a T-shaped space~\cite{Klinker1988} whose limbs represent the diffuse and specular components in the dichromatic reflection model. Approaches based on this model~\cite{Gershon1987,Klinker1990} assume that the scene has been segmented into different parts with a uniform diffuse distribution, which make them unpractical for highly textured scenes. 
More recent work alleviates the need for segmentation by integrating spatio-temporal information~\cite{Mallick2006spec}. Similar approaches with different color spaces such as HSI~\cite{Yang2013}, rotated RGB~\cite{Mallick2005}, or custom spaces~\cite{Bajcsy1996,Shen2013,Tan2004} have also been proposed.
As well as color-based techniques, the so-called \emph{neighborhood analysis} methods~\cite{Mallick2006dich,Tan2005,Yoon2006} leverage the information from neighboring pixels to infer the reflective component at a given point.
 
Multi-image approaches use strategies as varied as structured illumination~\cite{Lamond2009,Ma2007,Nayar2006}, different viewing angles~\cite{Jaklivc1993,Lin2002}, or light-field imaging~\cite{Meng2015,Wang2016}.
Another multiple image approach makes use of polarizing filters. With a single filter in front of the lens, Nayar~et al.~\cite{Nayar1993} recast the separation problem as a linear system using multiple captures under different polarizations. Building on their work, a number of methods propose to combine color information with polarizing filters to separate the reflective components~\cite{Ma2007,Kim2002,Nayar1997,Umeyama2004}.

Ma~et al.~\cite{Ma2007} and Debevec~et al.~\cite{Debevec2000} suggest the use of filters \emph{both} in front of the camera and the light sources, which leads to a slightly different model for the recorded intensity.

\begin{figure}[tb]
\centering
\includegraphics[width=0.7\linewidth]{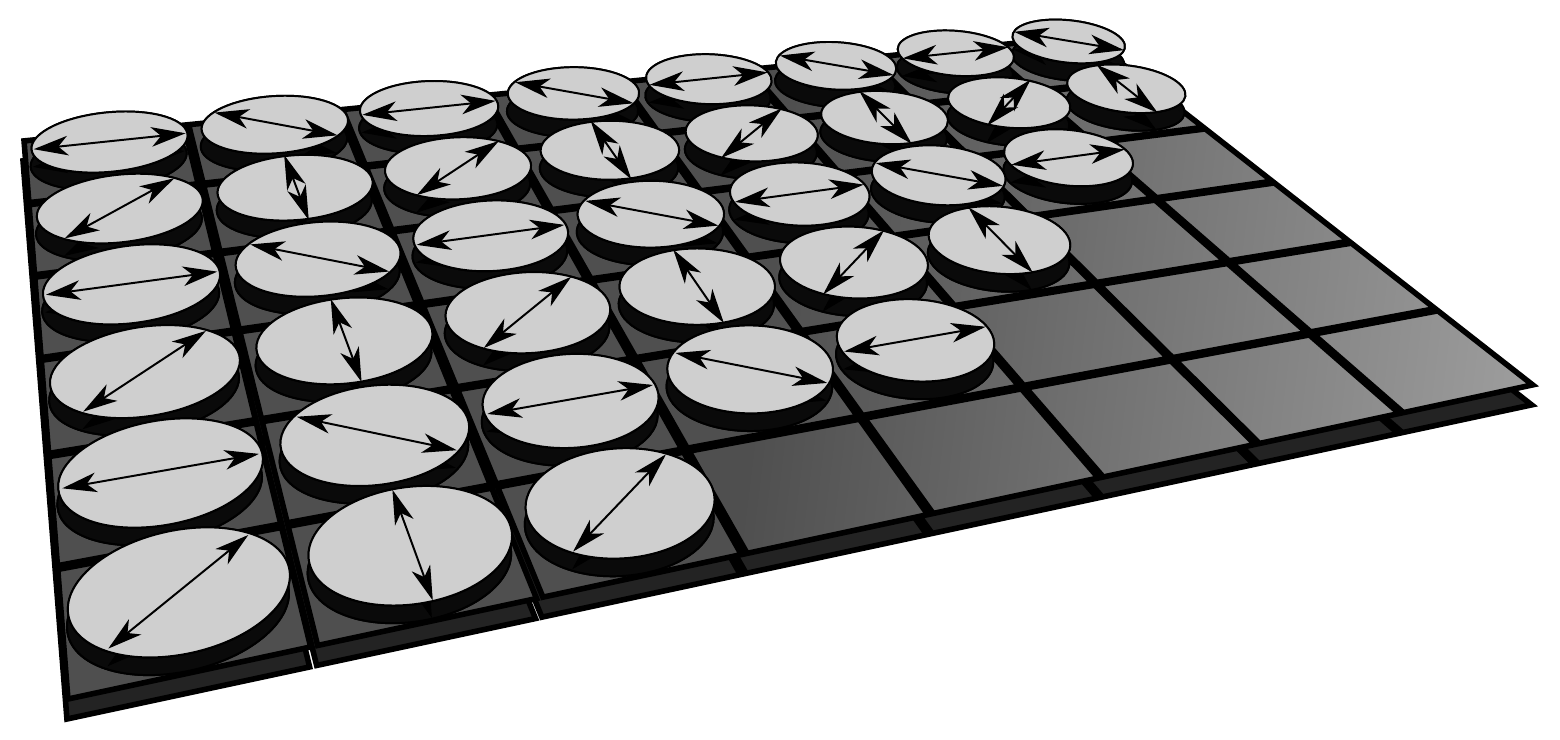}
\caption[]{Illustration of the concept of a filter array with different polarization orientations. The micro-filters are represented with disks and the arrow indicates the direction of polarization.
\label{fig:filter_array}
}
\end{figure}

Recently, cameras have been proposed with an array of polarizing micro-filters placed just in front of the image sensor~\cite{4DTechnologies,fluxdata,photonic-lattice,ricoh}. Analogously to a Bayer filter, the micro-filter of each pixel is oriented in one of a finite number of possible directions (see Fig.~\ref{fig:filter_array}).
Potentially, this can enable the separation of the diffuse and specular components from a single image with the accuracy and robustness of multi-image polarization methods.

\subsection{Proposed approach}
\label{proposed_approach}

In this paper, we investigate the feasibility of using these types of cameras for the separation of specularities by proposing an algorithm to demosaic images generated with polarizing micro-filter arrays. As well as regular patterns, we also study random arrangements with varying numbers of orientations and show that these arrangements offer advantages over regular patterns.


In addition to demosaicing, we demonstrate the usefulness of micro-filter arrays for photometric stereo. Using a light dome with several light sources having different polarizing orientations, we show that the separation of the diffuse and specular components using our design can significantly improve the estimation of the surface normals of objects. This approach is essentially the dual of~\cite{Cui2017}, where a camera with micro-polarizers is combined with a multi-view approach to infer the shape of objects.


The code and data required to reproduce all the results presented in this paper will be made available online with the camera-ready version.



\section{Problem statement}

Polarization is a physical property of light that characterizes the orientation of electromagnetic transverse waves in space.
A wave is said to be \emph{polarized} when its orientation follows a regular pattern in space: for example, when the wave oscillates in a fixed direction it is called \emph{linearly polarized}.
In contrast, \emph{unpolarized light} is composed of a mixture of electromagnetic waves with different orientations.
A \emph{linear polarizing filter} (or polarizer) is a physical device that only lets through light with a given polarization.
Given a light wave polarized in direction $\phi$ with initial intensity $I_0$, the polarizing filter will project it onto its orientation $\theta$. This is summarized by \emph{Malus' law}, which quantifies the light intensity $I$ after it goes through the filter: $I = I_0 \cos^2(\phi - \theta)$.
Unpolarized light can be thought of as containing all orientations equally, hence the resulting intensity is simply $I_0/2$, which is the mean value of $I_0 \cos^2(\phi-\theta)$ over all orientations. 


Polarization is of interest here because the diffuse and specular reflections affect polarization differently. Specular reflections are caused by direct reflection of the incoming light at an object's surface. Therefore, this type of reflection preserves the polarization of the incoming light.
Diffusion is slightly more complex. The diffusion created by the surface reflections can be thought of as the combination of several specular reflections in all orientations due to the rough nature of most materials at the microscopic level. The polarization of the light reflected by this microfacet model is not preserved. On the other hand, some of the light is also scattered inside the medium; the polarization of such light is retained.
We can leverage this observation to separate diffuse and specular reflection. In fact, we separate polarized and unpolarized light, which we can associate to the specular and diffuse components as long as the subsurface scattering is relatively weak.

In particular, suppose we have a scene illuminated by a single light source polarized with an angle $\phi$; this can be achieved by placing a linear polarizer in front of it. For now, assume that $\phi$ is known; later, we show that this is unnecessary as it can be easily estimated.
Furthermore, suppose we have a digital camera with a filter array of linear polarizers with $K$ different orientations in front of its sensor. We describe the image formation process as follows. Let $\vec{X}_k \in \R^{N \times N}$, $k = 1, 2, ..., K$, be a collection of $K$ images, each filtered with a distinct orientation $\theta_k$. From Malus' law, we can express each individual image as the sum of a constant diffuse term and a specular term that is modulated according to the polarization of the filters:
\begin{align}
\label{eq:diffuse_specular_matrix_form}
\vec{X}_k = \frac{1}{2}\vec{Z}_d + \vec{Z}_s \cos^2(\phi - \theta_k),
\end{align}
where $\vec{Z}_d$ and $\vec{Z}_s$ are the diffuse and specular components, respectively.
Equivalently, we can rewrite~\eqref{eq:diffuse_specular_matrix_form} by flattening $\vec{X}_k$ into a vector $\vec{x}_k = \vectorize(\vec{X}_k)$\footnote{Note that we use bold uppercase letters for matrix notation and the same letter in lowercase for its flattened version. Throughout this chapter, we interchangeably use both notations to denote the same object.}:
\begin{align}
\vec{x}_k &= \underbrace{\begin{bmatrix} \frac{1}{2}\vec{I} & \cos^2(\phi - \theta_k)\vec{I} \end{bmatrix} }_{\vec{C}_k}
\begin{bmatrix} \vec{z}_d\\
\vec{z}_s
\end{bmatrix}.
\end{align}
Here, $\vec{z}_d = \vectorize(\vec{Z}_d) \in \R^{N^2}$, $\vec{z}_s = \vectorize(\vec{Z}_s) \in \R^{N^2}$ and $\vec{C}_k \in \R^{N^2 \times 2N^2}$ is the matrix that modulates the specularity with the correct attenuation and combines the diffuse and specular terms.

To model a camera with a micro-array of polarizing filters, we need to select only one polarization orientation for each pixel. To model this, let $\vec{y}_k = \vec{A}_k\vec{x}_k$, where $\vec{A}_k \in \R^{N^2 \times N^2}$ is a mask that zeroes the measurements that we do not have access to; it has the form of a diagonal matrix with a $1$ where the pixel is selected and a $0$ otherwise. Note that, since we have one polarizing filter per pixel, $\sum_{k=1}^K \vec{A}_k = \vec{I}$.
At this stage, we make no assumption about the filter orientations: they can be regularly or irregularly structured over the array.
Finally, the measured image is given by
\begin{align}
\vec{y}  = \sum_{k=1}^K \vec{y}_k = \vec{AC} \begin{bmatrix}
\vec{z}_d\\
\vec{z}_s
\end{bmatrix},
\end{align}
where
\begin{align*}
\vec{A} &= [\vec{A}_1, \vec{A}_2, \dots, \vec{A}_K] \in \R^{N^2 \times KN^2}, \\
\vec{C} &= [\vec{C}_1, \vec{C}_2, \dots, \vec{C}_K]^\top  \in \R^{KN^2 \times 2N^2}.
\end{align*}

Using this formulation, our aim is to estimate $\vec{z}_d$ and $\vec{z}_s$ given the measurements $\vec{y}$ and the downsampling matrix $\vec{S} = \vec{AC}$. To simplify notation we write the ``superimage'' with both diffuse and specular components as $\vec{z}  = [\vec{z}_d, \vec{z}_s]^\top$.

\section{Algorithms}

We propose a general approach to separate the diffuse and specular components.
Finding both components using only a single image is an under-constrained problem, which we propose to regularize using the TV norm:
\begin{equation}
\begin{aligned}
\label{eq:direct_reconstruction}
\vec{\hat{z}}=&\argmin_{x}  \left\| \vec{y} - \vec{S} \vec{z} \right\|_2^2 + T(\vec{z}),
\end{aligned}
\end{equation}
where $T(\vec{z}) = \gamma_d \left\| \vec{Z}_d  \right\|_{TV}
+ \gamma_s \left\|\vec{Z}_s \right\|_{TV}$ is the regularization term. Here $\gamma_d$, $\gamma_s$ are regularization parameters and $\|\cdot\|_{TV}$ is the total variation (TV) norm\footnote{Technically, the TV norm is only a semi-norm.}. Typically, the TV norm is defined as the $\ell^1$ norm of the first order differential operator, although some authors also consider the $\ell^2$ norm of the same differential operator. A compromise between the two approaches is the \emph{Huber norm} or \emph{Huber loss}, defined as $L(\vec{z}) = \sum L(x_{i})$, where
\begin{align}
L(x) = \begin{cases}
\frac{1}{2}x^{2} & \mbox{for } |x| < 1\\
|x| - \frac{1}{2} & \mbox{otherwise.}
\end{cases}
\end{align}

We show below how to solve~\eqref{eq:direct_reconstruction} with these different variations of the TV norm.
The advantage of the $\ell^1$ approach is that, since it favors solutions that have a sparse first order derivative, it maintains edges in the image. In contrast, the $\ell^2$ approach can blur edges but performs well on other parts of the image. The $\ell^2$ formulation can also lead to faster algorithms.

\subsection{Minimizing the $\ell^2$ TV norm}

Although general optimization packages can be used to solve~\eqref{eq:direct_reconstruction}, the sampling matrix $\vec{S}$ is very large and therefore these techniques are limited to relatively small-sized images. In order to overcome this, we observe that~\eqref{eq:direct_reconstruction} can be expressed as a sparse linear system.
To see this, note that $T(\cdot)$, in the case of the $\ell^2$ TV norm, can be written as
\begin{equation}
\label{eq:tvnorm_matrix}
    T(\vec{z}) = \vec{z}^\top \vec{D}^\top \vec{W} \vec{D} \vec{z},
\end{equation}
where $\vec{D}$ is the $2$D discrete differential operator matrix and $\vec{W}$ encapsulates the effect of $\gamma_d$ and $\gamma_s$.
Finally, setting the derivative of~\eqref{eq:direct_reconstruction} to zero yields the following sparse linear system:
\begin{align}
\label{eq:derivative}
(\vec{S}^\top \vec{S} + \vec{D}^\top \vec{W} \vec{D}) \vec{\hat{z}} = \vec{S}^\top \vec{y}.
\end{align}
Although this system is very large, the matrix $\vec{S}^\top \vec{S} + \vec{D}^\top \vec{W} \vec{D}$ is sparse, symmetric and positive definite, and thus can be efficiently solved using approaches such as the conjugate gradient method. This leads to an algorithm that has linear complexity in the number of pixels.

\subsection{Minimizing the $\ell^1$ TV norm}

Unfortunately, unlike the $\ell^2$ case, we cannot find a closed form solution with the $\ell^1$ norm. 
Instead, we propose to use the split Bregmann method~\cite{Goldstein2009} to solve~\eqref{eq:direct_reconstruction} iteratively. In particular, \eqref{eq:direct_reconstruction} can be written as
\begin{equation}
\begin{aligned}
\label{eq:bregman_reconstruction}
&\vec{\hat{z}}= \argmin_{\vec{z}, \vec{d}} \left\| \vec{y} - \vec{S} \vec{z} \right\|_2^2 + \gamma_d \left\| \vec{d_d} \right\|_{1} + \gamma_s \left\| \vec{d_s} \right\|_{1}  \quad\text{s.t. }\quad\begin{bmatrix}
\vec{d_d} \\
\vec{d_s}
\end{bmatrix} = \vec{D}\begin{bmatrix}
\vec{z_d} \\
\vec{z_s}
\end{bmatrix},
\end{aligned}
\end{equation}
so that each step of the Bregmann iteration algorithm consists of solving
\begin{equation}
\label{eq:bregman_iteration}
    (\vec{z}^{k+1}, \vec{d}^{k+1}) = \argmin_{\vec{z}, \vec{d}} \left\| \vec{y} - \vec{S} \vec{z} \right\|_2^2 + \gamma_d \left\| \vec{d_d} \right\|_{1} + \gamma_s \left\| \vec{d_s} \right\|_{1} + \lambda \left\| \vec{d} - \vec{D} \vec{z} - \vec{b}^{k} \right\|_2^2,
\end{equation}
where $\vec{b}^{k+1} = \vec{b}^{k} + (\vec{D}\vec{z}^{k+1} - \vec{d}^{k+1})$. To solve~\eqref{eq:bregman_iteration}, we can alternate between minimizing over $\vec{z}$ and $\vec{d}$:
\begin{align}
& \vec{z}^{k+1} = \argmin_{\vec{z}} \left\| \vec{y} - \vec{S} \vec{z} \right\|_2^2 + \lambda \left\| \vec{d}^{k} - \vec{D} \vec{z} - \vec{b}^{k} \right\|_2^2, \label{eq:bregmann_x}\\
& \vec{d}^{k+1} = \argmin_{\vec{d}} \gamma_d \left\| \vec{d_d} \right\|_{1} + \gamma_s \left\| \vec{d_s} \right\|_{1} + \lambda \left\| \vec{d} - \vec{D} \vec{z}^{k+1} - \vec{b}^{k} \right\|_2^2\label{eq:bregmann_d}.
\end{align}

The main advantage of this approach is that \eqref{eq:bregmann_d} is now decoupled over space, and thus has a closed form solution. Additionally, \eqref{eq:bregmann_x} is now an $\ell^2$ minimization, which again has a closed form solution. In practice, only a few iterations of the split Bregmann algorithm are needed to converge so the method is acceptably fast and has linear complexity in the number of pixels in the image.

\subsection{Minimizing the Huber TV norm}

When using the Huber norm regularization, the minimization problem~\eqref{eq:direct_reconstruction} becomes
\begin{align}
\label{eq:huber_reconstruction}
&\vec{\hat{z}}= \argmin_{\vec{z}, \vec{d}} \left\| \vec{y} - \vec{S} \vec{z} \right\|_2^2 + \gamma_d L\left( \vec{D_s}\vec{z_d} \right) + \gamma_s L\left( \vec{D_s}\vec{z_s} \right).
\end{align}
Setting the derivative to 0, we obtain
\begin{equation}
\label{eq:huber_loss_diff}
2 \vec{S}^\top \vec{S} \vec{z} - 2 \vec{S}^\top \vec{y} + \gamma_d \vec{D_d}^\top L'\left( \vec{D_s}\vec{z_d} \right) + \gamma_s \vec{D_s}^\top L'\left( \vec{D_s}\vec{z_s} \right) = \vec{0}.
\end{equation}
As $L(\vec{z})$ is twice differentiable,~\eqref{eq:huber_loss_diff} is easy to solve using, for example, Newton's method. To initialize the iteration, we propose to use the solution of the $\ell^2$ TV norm.

\subsection{Unknown light polarization}

\begin{figure}[t]
\centering
\includegraphics[width=0.6\linewidth]{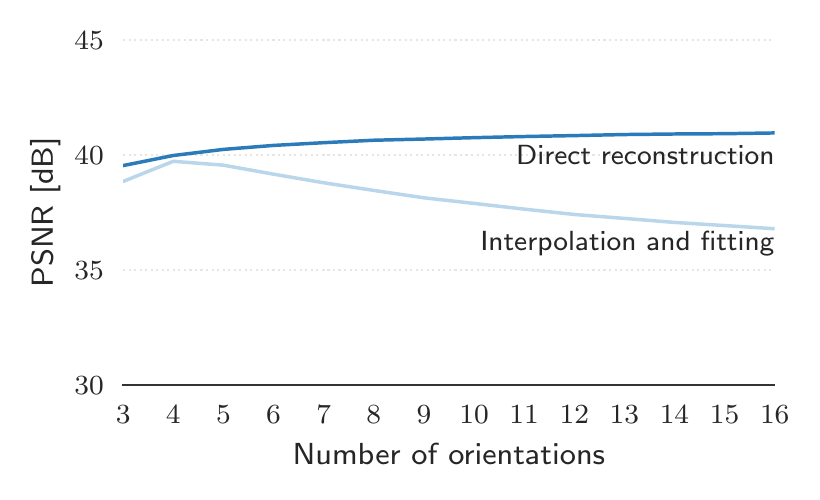}
\caption[]{{\redtext Influence of the number of orientations with a random filter design: comparison of our proposed algorithm, which performs the separation in one shot, with a two stage approach, which first interpolates the different channels and then extracts the specularity by fitting a cosine. Both methods use the $\ell^2$ TV-norm as regularizer.}
\label{fig:comparison_demosaicing}
}
\end{figure}

In the above description, we assumed that the polarization angle $\phi$ of the incoming light was known. Although it can be directly measured in controlled experiments, this is not necessary since it can be accurately estimated using the following procedure. 
First, for each image $\vec{z}_k$, we estimate its mean from the available pixels: \begin{equation}
    \mu_k = \sum_{n=1}^{N^2} [\vec{y}_k]_n / \sum_{n=1}^{N^2} [\vec{A}_k]_{nn}.
\end{equation}
Using the linearity of the mean, we have
\begin{equation}
    \begin{bmatrix}\mu_{1}\\
    \mu_{2}\\
    \vdots\\
    \mu_{K}
    \end{bmatrix} 
    =  \mu_d +  \begin{bmatrix} \cos(\phi - \theta_1)^2\\
    \cos(\phi - \theta_2)^2\\
    \vdots\\
    \cos(\phi - \theta_K)^2
    \end{bmatrix} \mu_s, 
\end{equation}
where $\mu_d$ and $\mu_s$ are the mean values of $\vec{z}_d$ and $\vec{z}_s$.
For $K > 3$, this set of non-linear equations is overdetermined, and finding $\phi$, $\mu_d$ and $\mu_s$ corresponds to solving the following minimization problem:
\begin{equation}
\begin{bmatrix}
\hat{\phi}\\ \hat{\mu}_d \\\hat{\mu}_s
\end{bmatrix}
= \argmin_{\phi, \mu_d, \mu_s} \sum_{k=1}^{K} \left(\mu_{k} - \mu_d - \mu_s \cos(\phi - \theta_k)^2\right)^2.
\end{equation}

{\redtext \section{Practical considerations}

Before evaluating and comparing our proposed approach to existing methods, we discuss below a few algorithmic and design choices.

 \begin{figure}[tb]
\centering
\includegraphics[width=\linewidth]{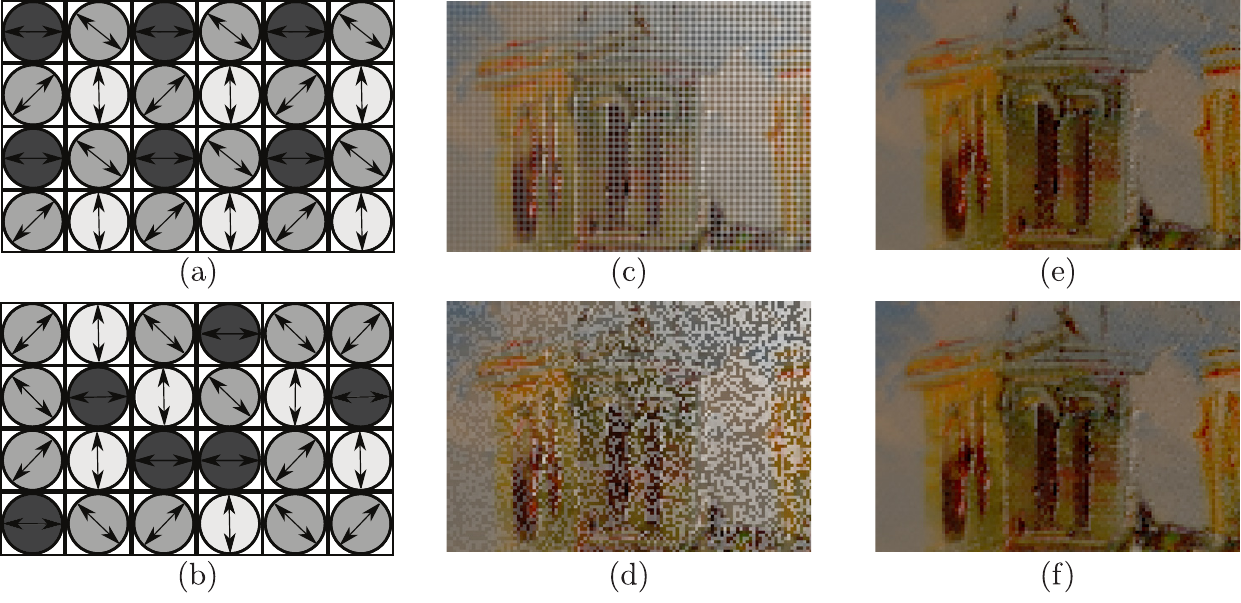}
\caption[]{Filter designs with $4$ different orientations: example of (a) a regular grid vs. (b) a random grid design.  Simulated image acquisitions using (c) a regular grid and (d) a random grid filter design. The last column illustrates the estimation of the diffuse component using our algorithm.
\label{fig:filter_design}
}
\end{figure}
 
\subsection{One-step vs. two-step algorithm}

Our proposed algorithm separates the components and interpolates the images in a single step. 
One might wonder how much we gain by performing this estimation in a single step as opposed to a two-stage approach where we first demosaic every individual image $\vec{X}_{k}$ and then fit a cosine to the interpolated images to identify the specular and diffuse components.

Figure~\ref{fig:comparison_demosaicing} shows a comparison between our direct reconstruction algorithm and the two-stage approach. Here, the same $\ell^2$ TV norm regularization is used for both the direct approach and the interpolation step of the two-stage approach.

Even though the two approaches are close in performance when the number of orientations is $4$, the two-stage approach performance clearly worsens when the number of orientations increases. Also, the two-stage approach peaks at $39.7$ dB, whereas the direct reconstruction continues to increase to $40.95$ dB.
While the observed performance gain is not huge, it is significant for many applications. 
The versatility of our algorithm over the two-stage approach is also an advantage when investigating different filter design patterns. In this regard, we observe that the direct reconstruction saturates at around $8$-$10$ orientations, suggesting that existing camera designs based on $4$ orientations are not optimal.
In the rest of this paper, we exclusively use the single-step approach.

}

{\redtext

 \begin{figure}[tb]
\centering
\includegraphics[width=\linewidth]{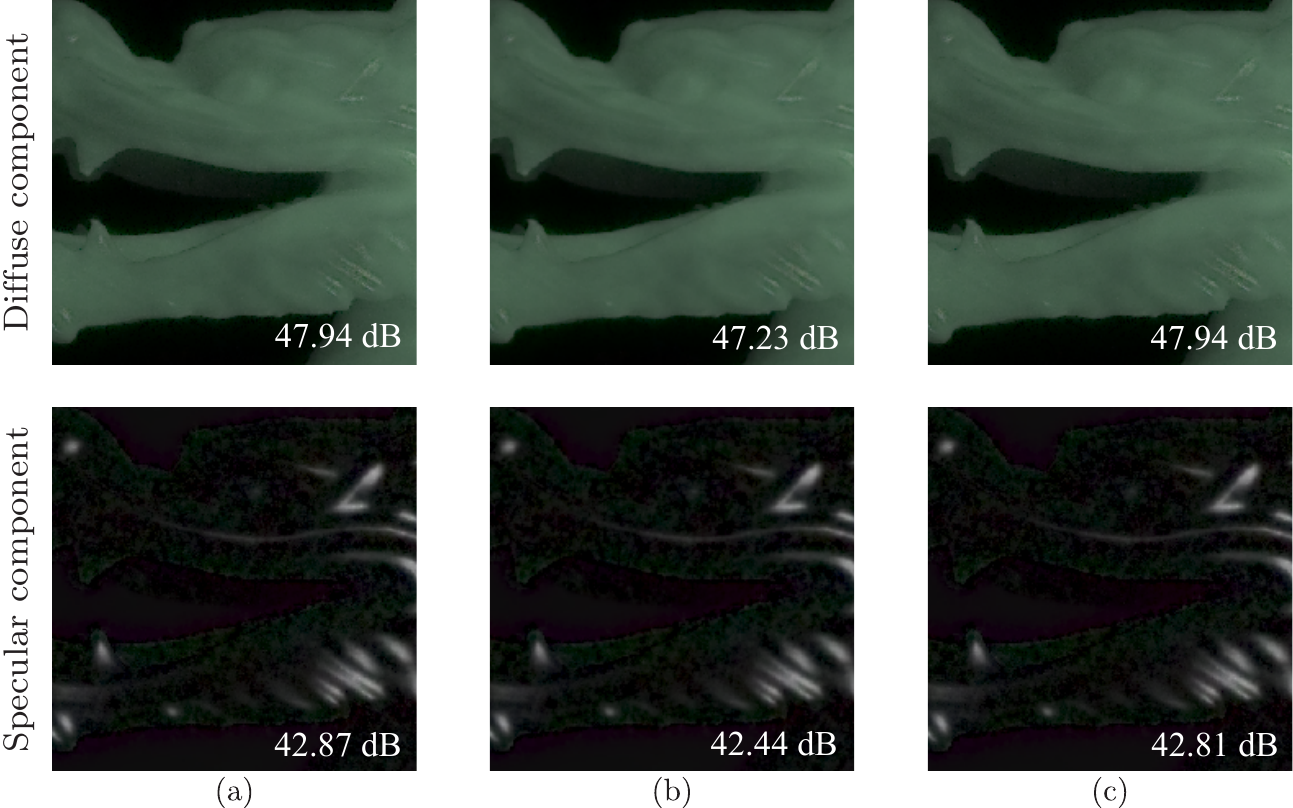}
\caption[]{Reconstructed diffuse and specular parts with (a) $\ell^2$, (b) $\ell^1$ and (c) Huber norms.
\label{fig:norm_comparison}
}
\end{figure}

\subsection{Comparison of the different regularizers}

As stated earlier, the TV norm is typically defined in terms of the $\ell^1$ norm, which leads to sharper edges, while the $\ell^2$ norm formulation is faster and can perform better on certain regions of the image. Finally, the Huber norm provides a tradeoff between the $\ell^1$ and $\ell^2$ variants. To quantify these differences, we ran a comparison of the results obtained with the three proposed norms. The results are given for one image in Fig.~\ref{fig:norm_comparison}. As we can see, there is not much difference in terms of PSNR, particularly between the $\ell^2$ and Huber norm regularization. We also observe that the $\ell^1$ regularized image is a bit sharper but also slightly noisier. In terms of performance, the $\ell^2$ and Huber norm algorithms are 5.7x and 3.3x faster than the $\ell^1$ algorithm, respectively. Based on these observations and computational speed, we choose to favor the $\ell^2$ norm for the rest of the experiments.
}

\subsection{Filter design patterns {\redtext and number of orientations}}

\begin{figure}[t]
 \centering
 \includegraphics[width=0.55\linewidth]{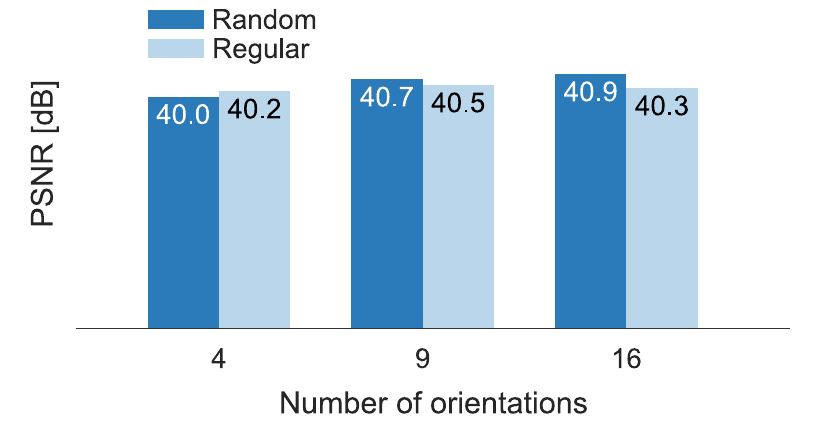}
 \caption[]{Comparison of the reconstruction error on the diffuse component with regular versus random grid designs.
 \label{fig:reg_vs_rand}
 }
 \end{figure}

There are a number of ways to design the micro-filter array, including (pseudo) random patterns and regular grids. Examples of images acquired with these designs are depicted in Fig.~\ref{fig:filter_design}c and~\ref{fig:filter_design}d. To obtain these images, we use Blender with the raytracing engine Cycles~\cite{Blender} and create ground truth diffuse and specular images from different render passes. We then simulate the effect of the filters by appropriately weighting and mixing the diffuse and specular components.
The corresponding estimated diffuse components are shown in Fig.~\ref{fig:filter_design}e and~\ref{fig:filter_design}f.
Additionally, in Fig.~\ref{fig:reg_vs_rand}, we compare the average performance of these two designs for a selection of synthetic images.
Overall, we observe that the filter design does not significantly influence the peak signal-to-noise ratio (PSNR). Nevertheless, as shown in Fig.~\ref{fig:filter_design}e and~\ref{fig:filter_design}f, random patterns lead to a slightly better qualitative estimation and, more importantly, the reconstruction artifacts induced by random patterns are more pleasing to the eye.

{\redtext \subsection{Cost of introducing the filter array}}

\begin{figure}[t]
\centering
\includegraphics[width=\linewidth]{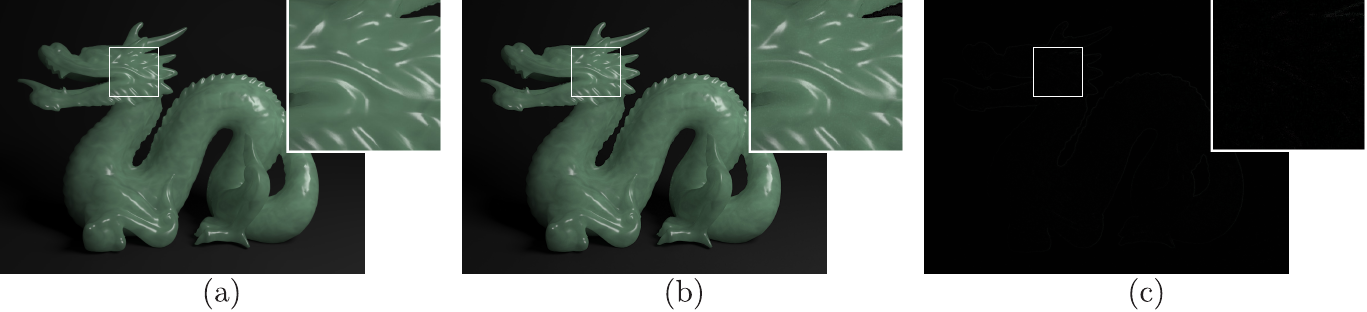}
\caption[]{Comparison of (a) the ground truth image to (b) the sum of our diffuse and specular estimations (PSNR = 43.8 dB). The difference (c) is almost zero suggesting that very little is lost by introducing a polarizing micro-filter.
\label{fig:consistency}
}
\end{figure}

 {\redtext Finally, }it is also interesting to investigate how much is lost by using a polarizing micro-filter, compared to a traditional camera, when one does not wish to separate the diffuse and specular components. In Fig.~\ref{fig:consistency}, we see that the sum of our diffuse and specular estimations is very close to the original image, suggesting that, even if the separation fails, the sum is almost indistinguishable from the output of a traditional camera.

\section{Experiments}

We test our {\redtext $\ell^2$  TV norm-based} algorithm in three different scenarios: rendered images, real images with simulated polarizing filters, and real images with real polarizers.
Simulated data is again generated with Blender Cycles raytracing engine~\cite{Blender}.
The second scenario consists of real images captured by Shen and Zheng~\cite{Shen2013}: the ground truth of the diffuse components is provided along with the original images.
We also provide our own captured images, taken under fixed polarized light sources, using a Nikon D810 DSLR camera with a polarizer in front of its lens. Note that even though the polarizing filter is not placed directly in front of the sensor, we neglect the effect that the lens might have on the light polarization. In practice, it may  slightly change the polarization phase, but this would be accounted for in the device calibration. To obtain a diffuse ground truth, we capture one image with the polarizing filter oriented orthogonal to the light's polarization.
For all setups, we first generate all complete images $\vec{X}_k$ for $k = 1, 2, \dots, K$ and then apply the downsampling operator $\vec{A}$ to simulate the effect of the polarized filter array.





\subsection{Single image diffuse and specular separation}
Given the results from Fig.~\ref{fig:reg_vs_rand}, we focus on a random filter array design and $16$ orientations for our algorithm. For all experiments, we set $\gamma_d = 0.01$ and $\gamma_s = 0.002$. It is possible to obtain slightly improved results with parameter tuning, but we avoided it.

To provide context, we compare our approach with the single image color-based technique proposed by Shen and Zheng in~\cite{Shen2013}. We should emphasize that their method takes as input a standard image while ours takes an image captured with the proposed filter array. Rather than being a fair comparison between algorithms, this experiment allows us to quantify the performance improvement offered by the proposed setup.

Performing this comparison raises a number of challenges in terms of color management. In particular, our algorithm operates in a linear color space, whereas~\cite{Shen2013} uses the sRGB space. To deal with this, we run our algorithm in the linear space and convert our estimated images to sRGB for comparison. Additionally, the images provided by~\cite{Shen2013} are in sRGB format and we convert them to a linear color space to apply our algorithm.

\begin{figure*}[htbp]
\centering
\includegraphics[width=1\linewidth]{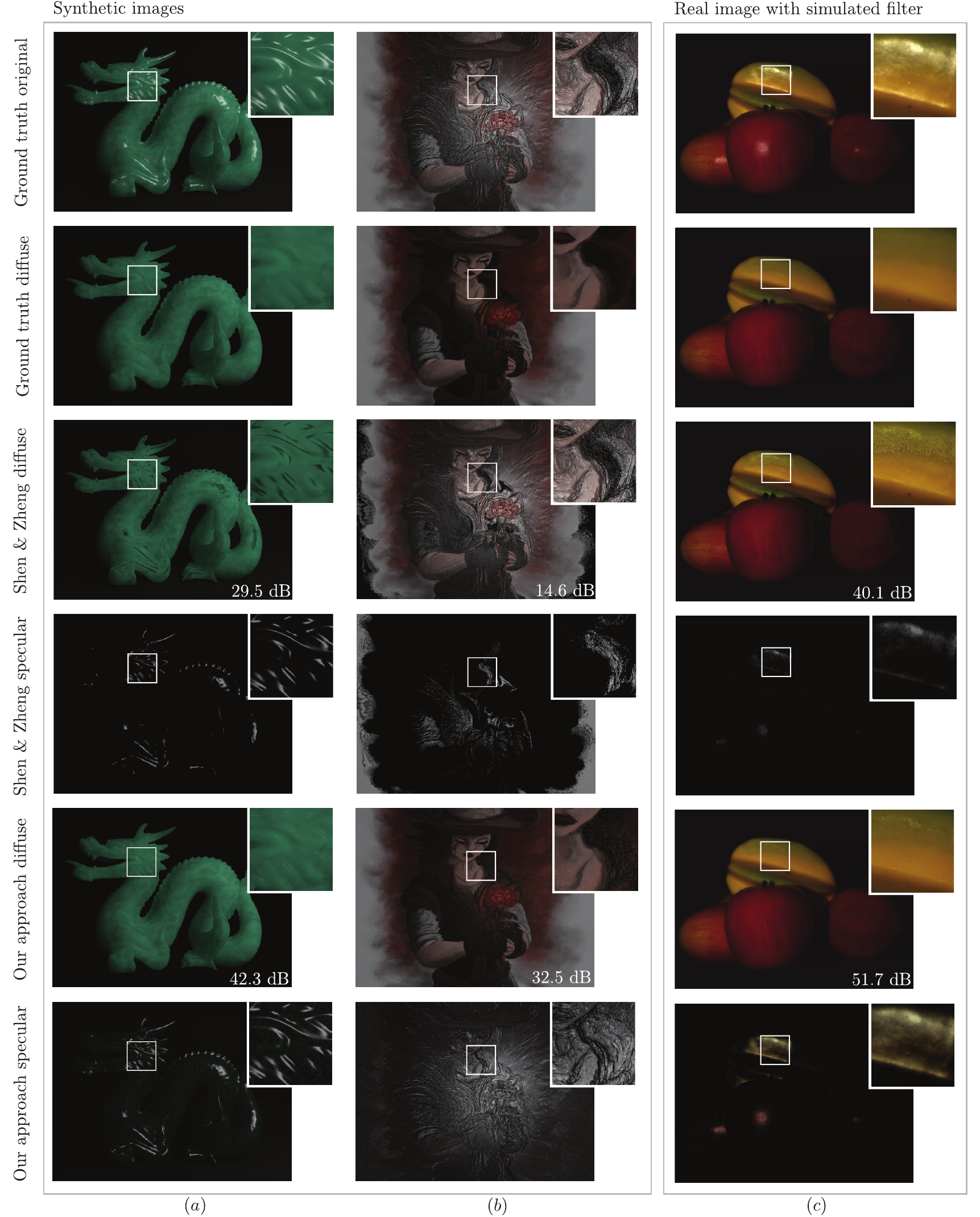}
\caption[]{Performance evaluation on two synthetic images and one real image with a simulated polarizer: (a) the Stanford dragon~\cite{StanfordScanRep}, (b) a digital painting with a computer-generated normal map, and (c) the fruits image from~\cite{Shen2013}.
\label{fig:results_1}
}
\end{figure*}

\begin{figure*}[htbp]
\centering
\includegraphics[width=1\linewidth]{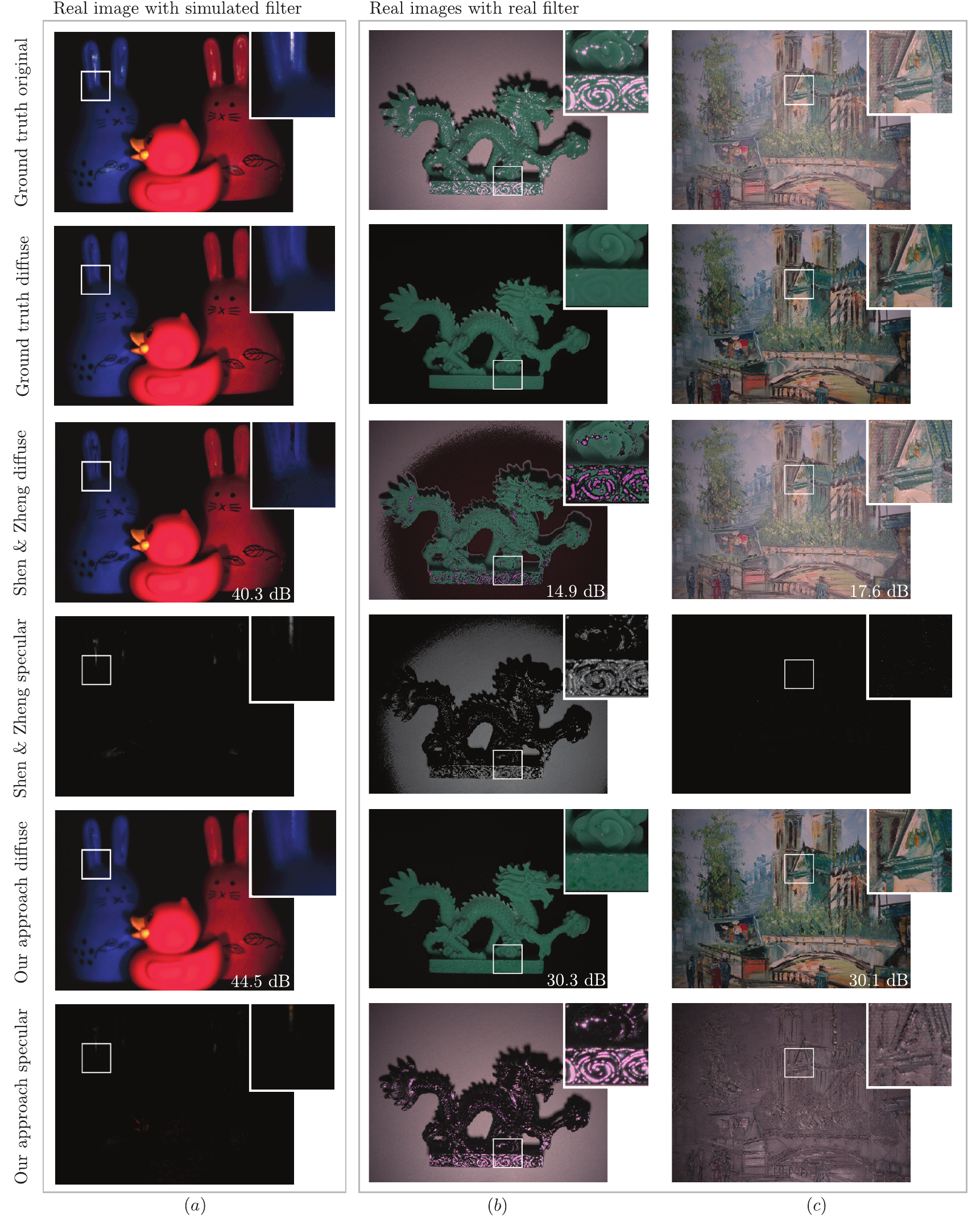}
\caption[]{Performance evaluation on three real images, one with a simulated polarizer and two with real polarizers: (a) the animals image from~\cite{Shen2013}, (b) a jade dragon, and (c) a painting.  
\label{fig:results_2}
}
\end{figure*}

Figures~\ref{fig:results_1} and~\ref{fig:results_2} depict the results for two images from each of the three different scenarios previously mentioned.
For all images, our custom setup significantly outperforms~\cite{Shen2013}. This is particularly true for images with complex surfaces such as Fig.~\ref{fig:results_1}b and~\ref{fig:results_2}b.

\begin{figure*}[!htb]
\centering
\includegraphics[width=1\linewidth]{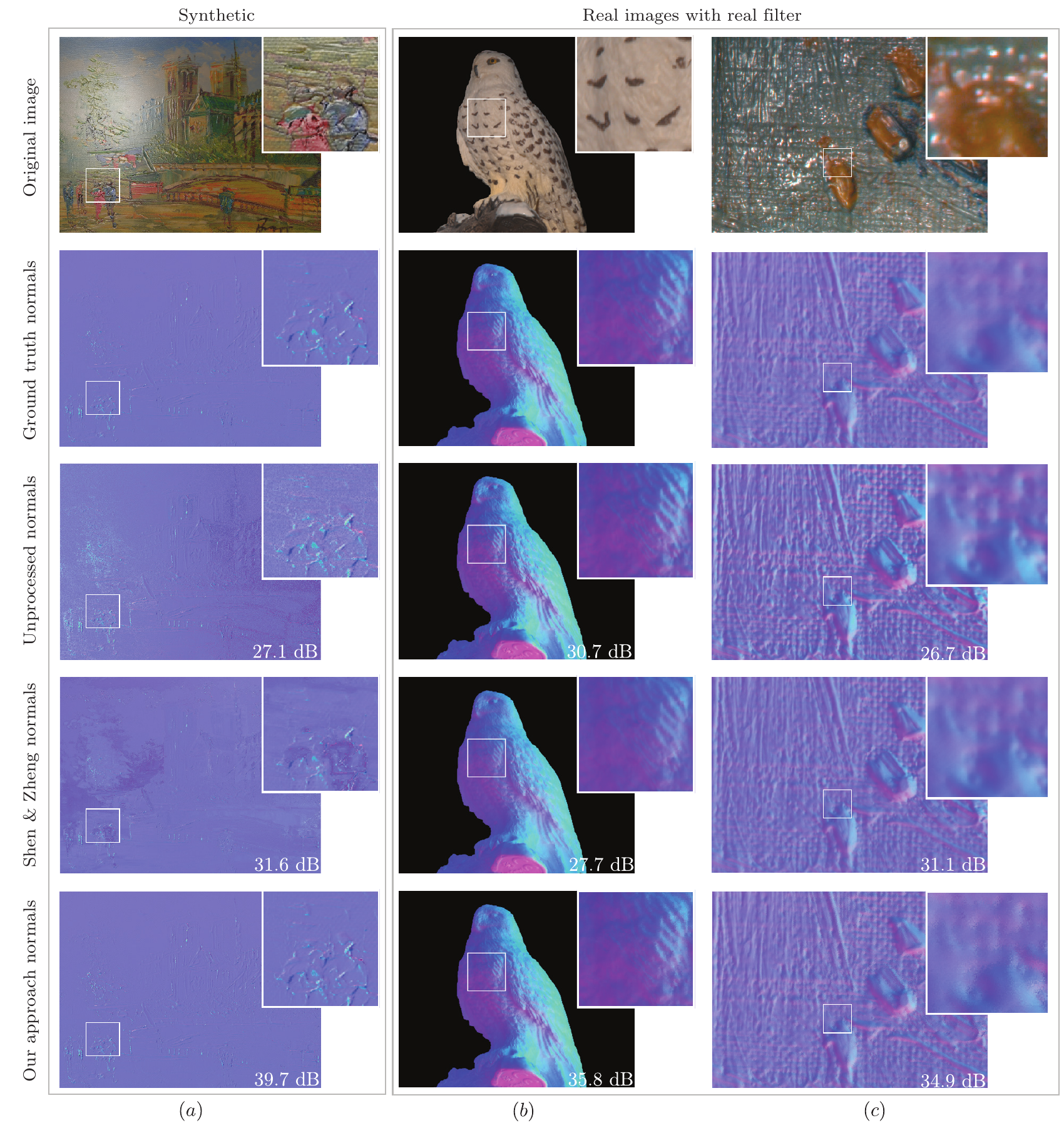}
\caption[]{Estimation of the normal map of three different scenes: (a) a rendered painting, (b) an owl figurine, and (c) a close-up of a real painting. From top to bottom, we show the original image, the ground truth normals, the normal estimation using the original image, the normal estimation using the proposed diffuse estimate and the normal estimation using the diffuse estimate from~\cite{Shen2013}.
\label{fig:results_dome}
}
\end{figure*}

\subsection{Photometric stereo}
 Photometric stereo~\cite{Woodham1980} infers surface normals from multiple images under different lighting by assuming that the surface of the objects are Lambertian materials. Unfortunately, this assumption is rarely satisfied as scenes often contain specular components and occlusions. In this section, we use our algorithm to first obtain a solid approximation of the diffuse part of the scene and then use it to recover the surface normals. 

For this experiment, we use a \emph{light dome} that is composed of $58$ lamps spread on a hemisphere surrounding the object of interest. A digital camera is placed at the zenith of the hemisphere.
We install linear polarizing filters with unknown orientation in front of each light source as well as in front of the camera. 
We capture several images, one under each illumination of the fixed lights for four different polarization orientations of the filter positioned in front of the camera; these four orientations are used to create mosaiced images on which we apply our algorithm to estimate the diffuse component. These diffuse images are then fed to the photometric stereo algorithm~\cite{Woodham1980}. In Fig.~\ref{fig:results_dome}, we compare the normal map generated by our algorithm with the normal map obtained from unprocessed images (i.e. with no separation of the diffuse component) as well as from Shen and Zheng's algorithm~\cite{Shen2013}. Note that the ground truth normal maps are computed using the non-mosaiced images. In all scenes, the gain of using our algorithm is clearly noticeable.

%% file: sections/conclusion.tex

\section{Conclusion}

We have studied the benefits of using a camera equipped with polarizing micro-filters for the separation of diffuse and specular terms. We presented a simple algorithm to demosaic images produced by such cameras and extract the diffuse term. 
Regarding the diffuse extraction, we have shown that our relatively simple algorithm can significantly outperform other single-image based techniques. A more accurate knowledge of the diffuse term can then be leveraged in other imaging applications such as photometric stereo; we demonstrate that our technique improves the estimation of the normal maps on various scenes.
For future work, we believe that including more elaborate priors based on, for example some of the existing color-based techniques, will improve the estimation.




%% file: ms.bbl
\begin{thebibliography}{10}

\bibitem{Woodham1980}
Woodham, R.J.:
\newblock Photometric method for determining surface orientation from multiple
  images.
\newblock Optical engineering \textbf{19} (1980)  191139

\bibitem{Artusi2011}
Artusi, A., Banterle, F., Chetverikov, D.:
\newblock A survey of specularity removal methods.
\newblock In: Computer Graphics Forum. Volume~30., Wiley Online Library (2011)
  2208--2230

\bibitem{Shafer1985}
Shafer, S.A.:
\newblock Using color to separate reflection components.
\newblock Color Research \& Application \textbf{10} (1985)  210--218

\bibitem{Klinker1988}
Klinker, G.J., Shafer, S.A., Kanade, T.:
\newblock The measurement of highlights in color images.
\newblock International Journal of Computer Vision \textbf{2} (1988)  7--32

\bibitem{Gershon1987}
Gershon, R.:
\newblock The Use of Color in Computational Vision.
\newblock Technical report. Department of Computer Science, University of
  Toronto (1987)

\bibitem{Klinker1990}
Klinker, G.J., Shafer, S.A., Kanade, T.:
\newblock A physical approach to color image understanding.
\newblock International Journal of Computer Vision \textbf{4} (1990)  7--38

\bibitem{Mallick2006spec}
Mallick, S.P., Zickler, T., Belhumeur, P.N., Kriegman, D.J.:
\newblock Specularity removal in images and videos: A {PDE} approach.
\newblock In: European Conference on Computer Vision, Springer (2006)  550--563

\bibitem{Yang2013}
Yang, J., Liu, L., Li, S.Z.:
\newblock Separating specular and diffuse reflection components in the {HSI}
  color space.
\newblock In: IEEE International Conference on Computer Vision Workshops.
  (2013)  891--898

\bibitem{Mallick2005}
Mallick, S.P., Zickler, T.E., Kriegman, D.J., Belhumeur, P.N.:
\newblock Beyond {L}ambert: Reconstructing specular surfaces using color.
\newblock In: IEEE Computer Society Conference on Computer Vision and Pattern
  Recognition. Volume~2., Ieee (2005)  619--626

\bibitem{Bajcsy1996}
Bajcsy, R., Lee, S.W., Leonardis, A.:
\newblock Detection of diffuse and specular interface reflections and
  inter-reflections by color image segmentation.
\newblock International Journal of Computer Vision \textbf{17} (1996)  241--272

\bibitem{Shen2013}
Shen, H.L., Zheng, Z.H.:
\newblock Real-time highlight removal using intensity ratio.
\newblock Applied optics \textbf{52} (2013)  4483--4493

\bibitem{Tan2004}
Tan, R.T., Nishino, K., Ikeuchi, K.:
\newblock Separating reflection components based on chromaticity and noise
  analysis.
\newblock IEEE Transactions on Pattern Analysis and Machine Intelligence
  \textbf{26} (2004)  1373--1379

\bibitem{Mallick2006dich}
Mallick, S.P., Zickler, T., Belhumeur, P., Kriegman, D.:
\newblock Dichromatic separation: specularity removal and editing.
\newblock In: ACM SIGGRAPH 2006 Sketches, ACM (2006)  166

\bibitem{Tan2005}
Tan, R.T., Ikeuchi, K.:
\newblock Separating reflection components of textured surfaces using a single
  image.
\newblock IEEE Transactions on Pattern Analysis and Machine Intelligence
  \textbf{27} (2005)  178--193

\bibitem{Yoon2006}
Yoon, K.J., Choi, Y., Kweon, I.S.:
\newblock Fast separation of reflection components using a
  specularity-invariant image representation.
\newblock In: IEEE International Conference on Image Processing, IEEE (2006)
  973--976

\bibitem{Lamond2009}
Lamond, B., Peers, P., Ghosh, A., Debevec, P.:
\newblock Image-based separation of diffuse and specular reflections using
  environmental structured illumination.
\newblock In: IEEE Computational Photography. (2009)

\bibitem{Ma2007}
Ma, W.C., Hawkins, T., Peers, P., Chabert, C.F., Weiss, M., Debevec, P.:
\newblock Rapid acquisition of specular and diffuse normal maps from polarized
  spherical gradient illumination.
\newblock In: Eurographics {Symposium} on {Rendering}, Eurographics Association
  (2007)  183--194

\bibitem{Nayar2006}
Nayar, S.K., Krishnan, G., Grossberg, M.D., Raskar, R.:
\newblock Fast separation of direct and global components of a scene using high
  frequency illumination.
\newblock In: ACM Transactions on Graphics. Volume~25., ACM (2006)  935--944

\bibitem{Jaklivc1993}
Jakli{\v{c}}, A., Solina, F.:
\newblock Separating diffuse and specular component of image irradiance by
  translating a camera.
\newblock In: International Conference on Computer Analysis of Images and
  Patterns, Springer (1993)  428--435

\bibitem{Lin2002}
Lin, S., Li, Y., Kang, S.B., Tong, X., Shum, H.Y.:
\newblock Diffuse-specular separation and depth recovery from image sequences.
\newblock In: European Conference on Computer Vision, Springer (2002)  210--224

\bibitem{Meng2015}
Meng, L., Lu, L., Bedard, N., Berkner, K.:
\newblock Single-shot specular surface reconstruction with gonio-plenoptic
  imaging.
\newblock In: Proceedings of the IEEE International Conference on Computer
  Vision. (2015)  3433--3441

\bibitem{Wang2016}
Wang, H., Xu, C., Wang, X., Zhang, Y., Peng, B.:
\newblock Light field imaging based accurate image specular highlight removal.
\newblock {PLOS ONE} \textbf{11} (2016)  e0156173

\bibitem{Nayar1993}
Nayar, S.K., Fang, X.S., Boult, T.:
\newblock Removal of specularities using color and polarization.
\newblock In: IEEE Computer Society Conference on Computer Vision and Pattern
  Recognition. (1993)  583--590

\bibitem{Kim2002}
Kim, D.W., Lin, S., Hong, K.S., Shum, H.Y.:
\newblock Variational specular separation using color and polarization.
\newblock In: MVA. (2002)  176--179

\bibitem{Nayar1997}
Nayar, S.K., Fang, X.S., Boult, T.:
\newblock Separation of reflection components using color and polarization.
\newblock International Journal of Computer Vision \textbf{21} (1997)  163--186

\bibitem{Umeyama2004}
Umeyama, S., Godin, G.:
\newblock Separation of diffuse and specular components of surface reflection
  by use of polarization and statistical analysis of images.
\newblock IEEE Transactions on Pattern Analysis and Machine Intelligence
  \textbf{26} (2004)  639--647

\bibitem{Debevec2000}
Debevec, P., Hawkins, T., Tchou, C., Duiker, H.P., Sarokin, W., Sagar, M.:
\newblock Acquiring the reflectance field of a human face.
\newblock In: SIGGRAPH. (2000)  145--156

\bibitem{4DTechnologies}
{4D Technologies}:
\newblock Polarization camera for image enhancement.
\newblock (\url{https://www.4dtechnology.com/products/polarimeters/polarcam})

\bibitem{fluxdata}
FluxData:
\newblock Polarization imaging camera {FD-1665P}.
\newblock (\url{http://www.fluxdata.com/products/fd-1665p})

\bibitem{photonic-lattice}
{Photonic Lattice}:
\newblock Polarization imaging camera {PI-110}.
\newblock
  (\url{https://www.photonic-lattice.com/en/products/polarization_camera/pi-110/})

\bibitem{ricoh}
{Ricoh {I}maging {C}ompany {Ltd}}:
\newblock Polarization camera.
\newblock (\url{https://www.ricoh.com/technology/tech/051_polarization.html})

\bibitem{Cui2017}
Cui, Z., Gu, J., Shi, B., Tan, P., Kautz, J.:
\newblock Polarimetric multi-view stereo.
\newblock In: Proceedings of the IEEE Conference on Computer Vision and Pattern
  Recognition. (2017)  1558--1567

\bibitem{Goldstein2009}
Goldstein, T., Osher, S.:
\newblock The split bregman method for l1-regularized problems.
\newblock SIAM J. Img. Sci. \textbf{2} (2009)  323--343

\bibitem{Blender}
{Blender Online Community}:
\newblock Blender - {A} 3D modelling and rendering package.
\newblock Blender Foundation, Blender Institute, Amsterdam. (2017)

\bibitem{StanfordScanRep}
{Stanford Graphics Lab}:
\newblock {The Stanford 3D Scanning Repository}.
\newblock (\url{http://graphics.stanford.edu/data/3Dscanrep/})

\end{thebibliography}
